\documentclass[10pt,journal,compsoc]{IEEEtran}
\ifCLASSOPTIONcompsoc
  \usepackage[nocompress]{cite}
\else
  \usepackage{cite}
\fi
\ifCLASSINFOpdf
\else
\fi
\usepackage[T1]{fontenc}

\usepackage{microtype}
\usepackage{graphicx}
\usepackage[export]{adjustbox}
\usepackage{xcolor}
\usepackage{subfigure}
\usepackage[inline]{enumitem}
\usepackage{booktabs} % for professional tables
\usepackage{multirow}
\usepackage{hyperref}

\usepackage{amsmath,amsfonts,amssymb,amsthm,dsfont}

\def\R{\mathbb{R}}
\def\P{\mathcal{P}}

\def\1{\mathds{1}}

\def\X{\mathcal{X}}
\def\Z{\mathcal{Z}}
\def\E{\mathcal{E}}

\def\D{\mathcal{D}}

\def\our{HyperColor}

\usepackage[noend]{algpseudocode}
\usepackage[plain]{algorithm}

\begin{document}

%\title{HyperColor: generation 3D color with colors}
%\title{hypernetwork Approach for Populating 3D Game Scenes with Auto-generated and Auto-colored 3D Objects}
\title{\our{}: A HyperNetwork Approach for Synthesizing Auto-colored 3D Models for\\Game Scenes Population}

\author{Ivan~Kostiuk,
        Przemysław~Stachura,
        Sławomir~K.~Tadeja,
        Tomasz~Trzciński~\IEEEmembership{Member,~IEEE,}
        and~Przemysław~Spurek% <-this % stops a space
%~\IEEEmembership{Member,~IEEE,}
\IEEEcompsocitemizethanks{\IEEEcompsocthanksitem I. Kostiuk, P. Stachura, P. Spurek are with Jagiellonian University, Faculty of Mathematics and Computer Science, Kraków, Poland; S.K. Tadeja is with Jagiellonian University, Institute of Applied Computer Science, Kraków, Poland; T. Trzciński is with Warsaw University of Technology, Faculty of Electronics and Information Technology, Warsaw, Poland. E-mail: przemyslaw.spurek@uj.edu.pl}
%\IEEEcompsocthanksitem Sławomir K. Tadeja is with Jagiellonian University, Faculty of Physics, Astronomy and Computer Science, Kraków, Poland}%

% \IEEEcompsocitemizethanks{\IEEEcompsocthanksitem M. Shell was with the Department
% of Electrical and Computer Engineering, Georgia Institute of Technology, Atlanta,
% GA, 30332.\protect\\
% % note need leading \protect in front of \\ to get a newline within \thanks as
% % \\ is fragile and will error, could use \hfil\break instead.
% E-mail: see http://www.michaelshell.org/contact.html
% \IEEEcompsocthanksitem J. Doe and J. Doe are with Anonymous University.}% <-this % stops an unwanted space
}


\markboth{}%
{}



\IEEEtitleabstractindextext{%
\begin{abstract}
%Designing a 3D game scene is a tedious task that often requires a substantial amount of work. Typically, this task involves synthesis, coloring, and placement of 3D models within the game scene. To lessen this workload, we can apply machine learning and other procedural methods to automate some aspects of the game scene development. Earlier research has already tackled generating 3D models as well as automated generation of the game scene background with machine learning. However, model auto-coloring remains an underexplored problem. The automatic coloring of a 3D model is a challenging task, especially when dealing with the digital representation of a colorful, multipart object. In such a case, we have to ``understand'' the object's composition and coloring scheme of each part. Existing methods attempt to overcome this problem by relying on computer-aided design or photogrammetry methods, or separately color elements of the model. However, these techniques have their own caveats such as the need for segmentation of the object or generating individual parts that have to be assembled together to yield the final model which increases resulting mesh complexity. To this end, we address these limitations by training the deep learning model to automatically color the 3D models without the need to ``understand'' its composition to individual parts.

Designing a 3D game scene is a tedious task that often requires a substantial amount of work. Typically, this task involves synthesis, coloring, and placement of 3D models within the game scene. To lessen this workload, we can apply machine learning to automate some aspects of the game scene development. Earlier research has already tackled  automated generation of the game scene background with machine learning. However, model auto-coloring remains an underexplored problem. The automatic coloring of a 3D model is a challenging task, especially when dealing with the digital representation of a colorful, multipart object. In such a case, we have to ``understand'' the object's composition and coloring scheme of each part. Existing single-stage methods have their own caveats such as the need for segmentation of the object or generating individual parts that have to be assembled together to yield the final model. We address these limitations by proposing a two-stage training approach to synthesize auto-colored 3D models. In the first stage, we obtain a 3D point cloud representing a 3D object, whilst in the second stage, we assign colors to points within such cloud. Next, by leveraging the so-called triangulation trick, we generate a 3D mesh in which the surfaces are colored based on interpolation of colored points representing vertices of a given mesh triangle. This approach allows us to generate a smooth coloring scheme. Experimental evaluation shows that our two-stage approach gives better results in terms of shape reconstruction and coloring when compared to traditional single-stage techniques.

% \tomek{This abstract motivates the problem and enumerates the limitations of the existing methods, but it fails to deliver a description of what the clue of the proposed method is.2} \slawek{I changed the ending and remove a sentence about photogrammetry.}
%\tomek{We still need to add sthg like: Our approach outperforms the existing methods in an extensive quantitative/qualitative evaluation.}\slawek{Added last sentence.}
\end{abstract}

}

\maketitle

%\IEEEdisplaynontitleabstractindextext

% \fi
%\IEEEpeerreviewmaketitle

\IEEEraisesectionheading{\section{Introduction}\label{sec:introduction}}


\IEEEPARstart{O}{ne} of the most tedious and mundane tasks in the 3D game development is the preparation of the 3D scenes for individual game levels \cite{procedular_landscape,procedural_dungeons,indoor_procedural_lvl}. Often these levels are similar to each other because they are a modified versions of a previous ones~\cite{procedular_landscape, procedural_dungeons, indoor_procedural_lvl}. The scene content is typically composed of the background and adequately colored 3D models populating the given scene \cite{procedular_landscape}. Moreover, the population task usually imposes certain constraints, as the 3D objects have to be placed in a chosen spatial relation to each other and the terrain \cite{8382283, procedular_landscape}. Hence, we can assume that the scene creation process has to follow certain rules contingent on the game genre and particular level design \cite{8382283, procedural_dungeons}.

One of the ways to reduce the burden of the level design is to automate the process of scene content creation with procedural and machine learning (ML) methods \cite{arshad2020progressive, 8382283, procedular_landscape, procedural_terrain_overview, indoor_procedural_lvl, procedural_dungeons, GAN_lvl_3D, Rosenberg}. Besides the terrain creation which can be achieved separately, we can consider the automatic content creation process as a series of steps. These steps include: (i) generating plausible and diversified 3D models, (ii) coloring or overlaying these models with textures, and (iii) their proper placement in the 3D scene.
% \begin{enumerate}[label=(\roman*)]
%     \item creating the terrain;
%     \item generating plausible and diversified 3D models;
%     \item coloring or overlaying these models with textures;
%     \item their proper placement in the 3D scene background.
% \end{enumerate}
Some of these steps are already addressed in the literature, e.g. (i) terrain creation \cite{procedular_landscape, procedural_terrain_overview, indoor_procedural_lvl, procedural_dungeons}, or (ii) 3D model generation \cite{Wu20153DSA,Richter2015DiscriminativeSF,Tatarchenko2017OctreeGN,8100184,girdhar2016learning,10.5555/3157096.3157106,7298807,Choy20163DR2N2AU,Richter2018MatryoshkaNP,spurek2020hyperflow}. While the others, e.g. automatic model placement \cite{8382283, procedural_terrain_overview}, and automatic coloring \cite{textures_gen}, remain underexplored problems \cite{arshad2020progressive}. 

\begin{figure*}[th!]
\begin{center} 
 \includegraphics[width=\linewidth]{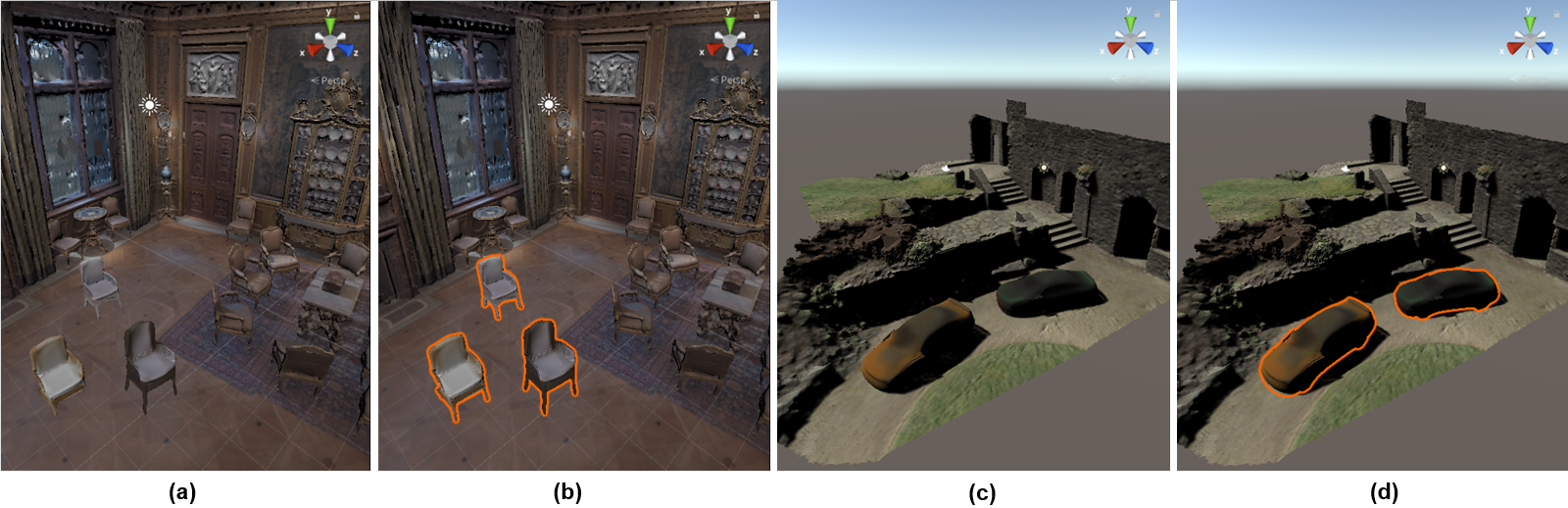}
\end{center} 
\caption{Hypernetwork generated and auto-colored chair and car 3D models populating game scenes. (a-b) The scene background was constructed out of photogrammetric model of \textit{the Great Drawing Room} at The Hallwyl Museum. Whereas (c-d) backgrounds is a photogrammetric model of a landscape (see \url{https://sketchfab.com/noxfcna}). Both backgrounds were downloaded from \cite{Maggiordomo2020RealWorldTT} and are presented in the editor of Unity game engine.}
\label{fig:gamescenesChairsCars} 
\end{figure*}

In this context, in this work, we are tackling the (i) and (ii) steps of the automatic content creation process. We propose a merger of these two steps by providing a method for simultaneous synthesis and auto-coloring of diversified 3D models \cite{arshad2020progressive}. This task requires a high level of ``understanding'' of what is being colored, i.e., the knowledge of what type of an object we are dealing with, including its real-world surfaces coloring and texturing. The coloring task becomes even more challenging when the object in question is composed of several parts requiring different coloring schemes \cite{arshad2020progressive}.

Existing methods  attempt to remedy this issue by segmenting the 3D object \cite{Mo_2019_CVPR} and training the neural network to single-color or texture individual object's parts. In such a case, the resulting mesh would have a more complex structure than needed or require additional mesh assembly procedures. %Recently, there were some successful attempts in generating dense, colored point clouds \cite{arshad2020progressive}. 

%Existing methods that try to remedy this issue rely on using computer-aided design (CAD) methods where the model is prepared with automatedly or manually assigned to given part color or texture. Alternative approach is the creation of photo-realistic 3D models using, for instance, the photogrammetry method \cite{thompson_manual_1966,tadeja_exploring_2021}. In the latter case, models are generated from a set of images from which their 3D shape and texturing are being automatically extracted. However, this process is time and resource-consuming. It also requires access to a set of real-life objects that needs to be digitally scanned. Yet another approach is to segment the 3D object \cite{Mo_2019_CVPR} and train the neural network to single-color or texture individual object parts. Here, the resulting mesh would have more complex structure than needed or require additional mesh assembly procedures. Recently, there were some successful attempts in generating dense, colored point-clouds \cite{arshad2020progressive}.

%automatic recognition, i.e., segmentation of the objects' parts and their individual coloring (or texturing). 
%Some of the earlier research proposes to use \textit{Generative Adversarial Networks} (GANs) for the procedural synthesis of textures \cite{textures_gen}. Whereas \cite{7536161} applied 

\begin{figure*}[th!]
\begin{center} 
 \includegraphics[width=\linewidth]{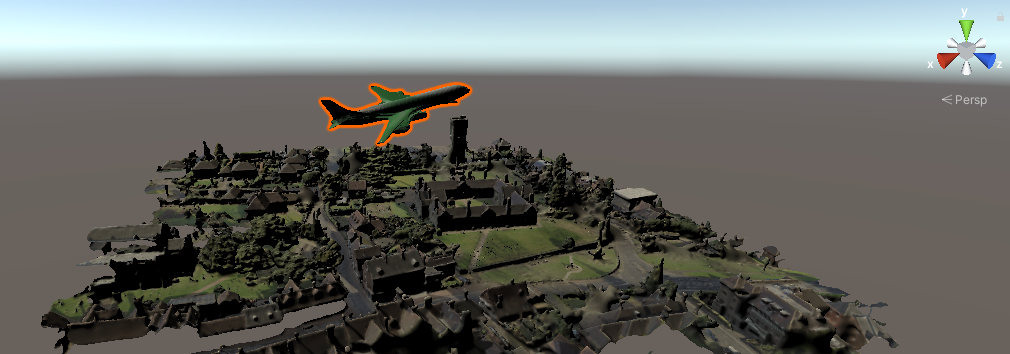}
\end{center} 
\caption{The scene background was constructed out of photogrammetric model made by the \textit{Aircam Surveys Ltd.} and downloaded through \cite{Maggiordomo2020RealWorldTT}.}% \slawek{Zastanawiam sie czy nie przerzucic tych dwoch pierwszych ilustracji ze scenami na koniec? Wtedy mamy powoli budowany obraz: chmury punktow->modele->sceny. Z drugiej strony przyciagaja uwage czytelnika od poczatku. Jak uwazacie?} \przemek{ja bym zostawił tutaj}}
\label{fig:gamescenesPlane} 
\end{figure*}

In this work, we are reformulating the problem of simultaneous generation and auto-coloring of 3D models in terms of deep neural network optimization. More precisely, we train a deep learning model to automatically color the 3D models without the need of ``understanding'' its decomposition to individual parts.
%However, this problem can be reformulated in a different way. We can train the deep learning model to automatically color the 3D models without the need of ``understanding'' its composition to individual parts.\tomek{these are the real probems we need to face and they should be elaborated.}
%To this end, we are proposing the use of \textit{HyperColor} machine learning model.
% \tomek{Let me rewrite what I understood so far and maybe let's use this as a backbone of the next draft of the intro:
% 1. Game scene design is a tedious task that requires a lot of work.
% 2. People can use some ml to automate that
% 3. Some works already started to do so, e.g. generating plausible and diversified 3D models, e.g. by solving for mesh-based point cloud completion ~\cite{spurek2020hyperflow}, yet there are some missing steps, e.g. coloring.
% 4. coloring of 3D is tough, however, because you need to understand what you are coloring and how (semantics, scene geometry etc). Existing methods attempt to do this this way ... but there are limitations of these approaches (which exactly?)
% 5. In this work, we address those limitations by reformulating the problem: we can train the deep learning model to automatically color the 3D models without the need of ``understanding'' its composition to individual parts.}
Our method \our{}\footnote{We make our implementation available at \url{https://github.com/KostiukIvan/HyperColor} }  allows for auto-generation and auto-coloring of diversified 3D models. We use a two-stage training process to obtain colored 3D point clouds (see Fig.~\ref{fig_rec:point:col}) which is later converted into 3D meshes by leveraging a \textit{triangulation trick}  \cite{spurek2020hypernetwork} (see Fig.~\ref{fig:chair_sphere_triangles}). These 3D objects can be incorporated into the 3D background resulting in matching and plausible game scenes (see Fig.~\ref{fig:gamescenesChairsCars}--\ref{fig:gamescenesPlane}).

%\tomek{I'm missing some information about the fact that our approach is better than the competitors. If we have space (and I believe we do) we can also give the overview of the rest of the paper: In the remainder of this work, we...}

In the remainder of this work, we first offer the relevant literature. Next, we will describe in detail our two-stage approach to generating plausible 3D auto-colored meshes using hypernetworks. Lastly, we will present the results of a comparison between our two-stage method against the traditional one-stage approach (see Tab.~\ref{tab:1}).

% In the remainder of this work, we first present to date research efforts regarding the ML-based approach to generating 3D objects as well as the coloring of such 3D models. Moreover, we briefly describe a procedural approach to game scene generation.

% Next, we will describe in detail our two-stage approach to generating plausible 3D auto-colored meshes using hypernetworks. Here, we first train an ML model to generate a 3D point cloud representing a given object and color them in the next stage by the means of a different deep learning model. Then, we transform the colored point cloud into a 3D mesh colors of which are interpolated out of the points constituting the used point cloud.

% Finally, we offer the results of comparison between three different classes of models (i.e. airplanes, cars and chairs) generating with our two-stage method against the traditional one-stage approach. The outcome of this study shows that the 3D objects generated by us are better in terms of both reconstructions as well as coloring (see Tab.~\ref{tab:1}). 

\section{Related Work}
We divide the related work into three parts. First, we describe the existing methods for generating 3D objects. Second,  we offer a discussion of models producing colors for 3D objects. Finally, to provide a wider context, we present a brief description of procedural methods used in generating game scenes and entire levels. 

\subsection{Generating 3D Objects}
Point clouds are among the most popular digital representations of 3D objects, widely used in LIDARs and depth cameras. Complex perception tasks that use point clouds, such as localization or object recognition, typically treat these representations as fixed-dimensional matrices, which requires subsampling of the cloud or other pre-processing~\cite{achlioptas2018learning,gadelha2018multiresolution,zamorski2018adversarial,sun2020pointgrow,yifan2019patch} .% involves ne They are used by many space mapping and registration devices, such as LIDARs and depth cameras. Whe\slawek{Brakuje mi tutaj chyba referencji do poprzedniego zdania.} Historically, generative models represent point clouds as a fixed-dimensional matrix (i.e., $N\times3$ where $N$ is a fixed number of points) \cite{achlioptas2018learning,gadelha2018multiresolution,zamorski2018adversarial,sun2020pointgrow,yifan2019patch} 
In practice, such an approach is often very limiting as the complexity of point clouds varies across object types and some objects need more points to represent their details than the others. %. On the other hand, we have many simpler real-life objects, that do not require as many data points to be exhaustively described. 

%A static number of points is a significant limitation for some applications. In practice, some objects are more complex and need more points to describe all details. On the other hand, we have many simple objects, which do not require as many elements. 

To remedy this issue, PointFlow \cite{yang2019pointflow} proposes to learn a two-level hierarchy of distributions where the first level is the distribution of shapes, while the second is the distribution of points from the object's surface. This formulation allows to sample an arbitrary number of points from a given shape. Moreover, PointFlow \cite{yang2019pointflow} applies normalizing flow \cite{papamakarios2021normalizing} to model surface of the object.

In \cite{spurek2020hypernetwork,spurek2020hyperflow} the authors propose to use a hypernetwork architecture to model distribution of shapes. Instead of producing a fixed number of points, hypernetwork generate many neural networks, a single network per object. Such models take an element from the prior distribution and transfer it to the object's surface. 
Such an approach allows to produce as many points as we need. We can sample an arbitrary number of points and transfer them using a target network. 

In another work~\cite{cai2020learning}, the authors propose to model a shape by learning the gradient field of its log-density. \cite{sun2020pointgrow} uses auto-regressive architecture
to model the distribution of 3D points, while in \cite{li2018point} the authors leverage a GAN architecture for the same purpose. All the above methods work on raw 3D point clouds without any colors.

% 3D model generation using ML
% \cite{Wu20153DSA,Richter2015DiscriminativeSF,Tatarchenko2017OctreeGN,8100184,girdhar2016learning,10.5555/3157096.3157106,7298807,Choy20163DR2N2AU,Richter2018MatryoshkaNP,spurek2020hyperflow}

\subsection{3D Objects with Colors}
Generation of point clouds with colors is explored to a much lesser extent than shape generation. Traditional colorization methods rely on human input \cite{huang2005adaptive,ironi2005colorization,luan2007natural}, which is time consuming and requires high level of expertise.

To solve for this problem, recent works automate the task of coloring with machine learning. In \cite{cao2019point,liu2019pccn} the authors present methods based on the pix2pix \cite{isola2017image} pipeline using cGAN. The generator attempts to predict the color of each point using PointNet \cite{qi2017pointnet}, and the PointNet-based discriminator attempts to judge fake color from the generator or real ground truth color.

\cite{arshad2020progressive} presents a conditional generative adversarial network that creates dense 3D point clouds with colors. A point transformer progressively grows the network. Every training iteration evolves a point vector into a point cloud of increasing resolution. This model can produce shapes with colors, however, its performance is evaluated only using the quality of shapes. 

In \cite{shinohara2021point2color} the authors use a conditional generative adversarial network (cGAN).
To achieve 3D point cloud colorization, the colors are estimated by PointNet++ \cite{qi2017pointnet++} and rendered into 2D images. The network is then trained by minimizing the distance between the real color and a colorized fake color.

All the above methods solve the problem of colorizing 3D point cloud. In this paper, we introduce a generative model that produces 3D point clouds with colors. Furthermore, we can create meshes with colors.  

%%%%%%%%%%%%%%%%%%%%%%%%%%%%%%%%%%%%%%%%%%%%%%%%%%5
\subsection{Procedural Game Scenes Generation}
The idea of using procedural tools to generate game scenes or entire levels is not new \cite{procedular_landscape, procedural_terrain_overview, indoor_procedural_lvl, procedural_dungeons, GAN_lvl_3D, Rosenberg}. These methods were used to generate indoor (e.g. \cite{indoor_procedural_lvl, procedural_dungeons}) and outdoor (e.g. \cite{procedular_landscape, procedural_terrain_overview}) environments alike. As a generation of the game levels is not the main goal of our work, we offer only a small selection of such earlier research carried out by other authors.

In the case of the outdoor level-generation method, the authors of \cite{procedular_landscape} presented a method for generating large-scale neutral game levels using natural systems. The substantial advantage of using this approach is the ability to edit the resulting scene after its initial generation.
A brief overview of procedural terrain generation can be found in \cite{procedural_terrain_overview}.

\begin{figure}
\begin{center} 
 \includegraphics[width=0.91\linewidth]{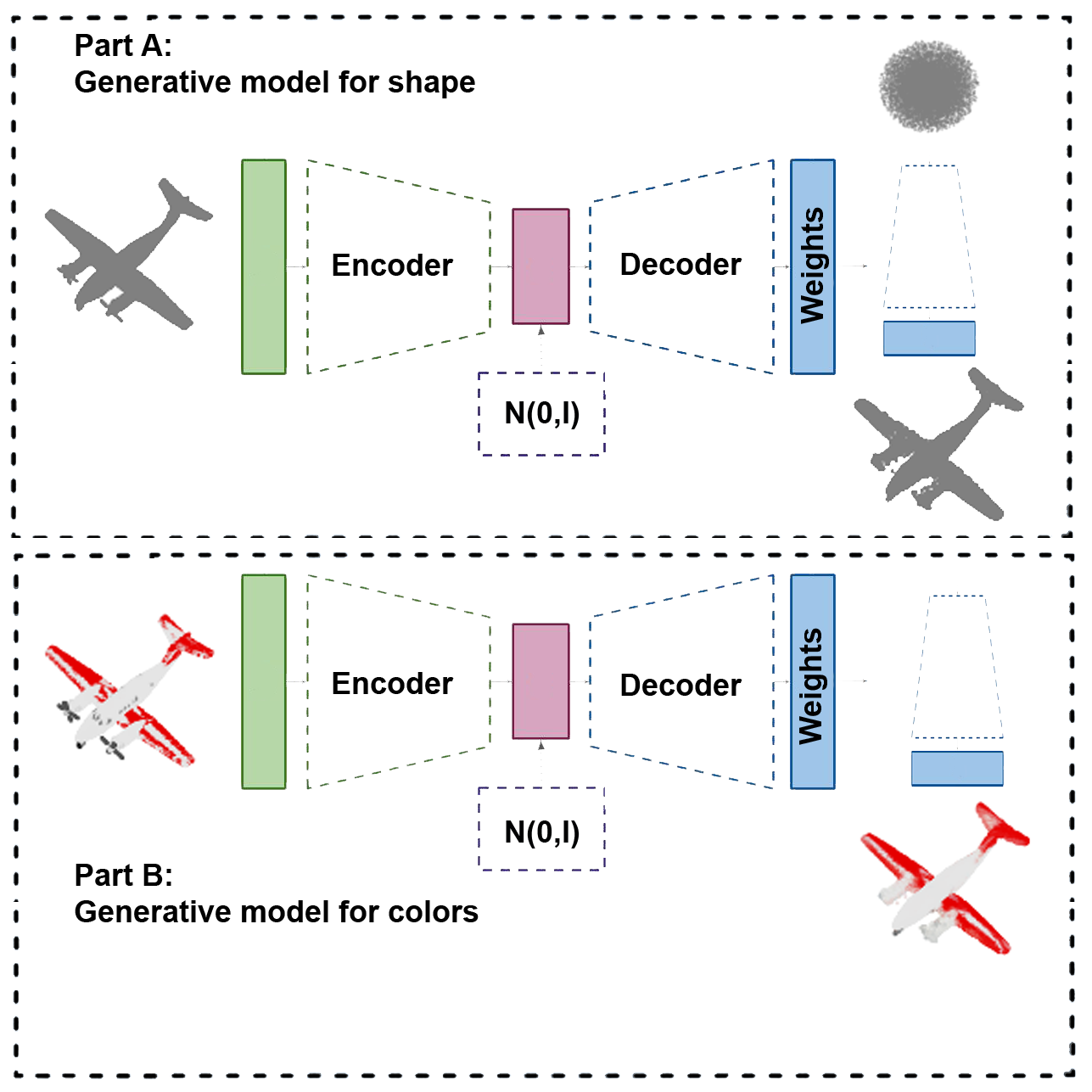}
\end{center} 
\caption{Our \our{} model consists from two parts. In the firs step we train classical model (HyperCloud) for shape reconstruction, see Part A. In the second one we train model to generate colors for existing 3D objects. Such solution allows to producing many different objects with many different colors.} 
\label{fig:model} 
\end{figure}

\begin{figure*}
\begin{center} 
    \includegraphics[width=0.9\linewidth]{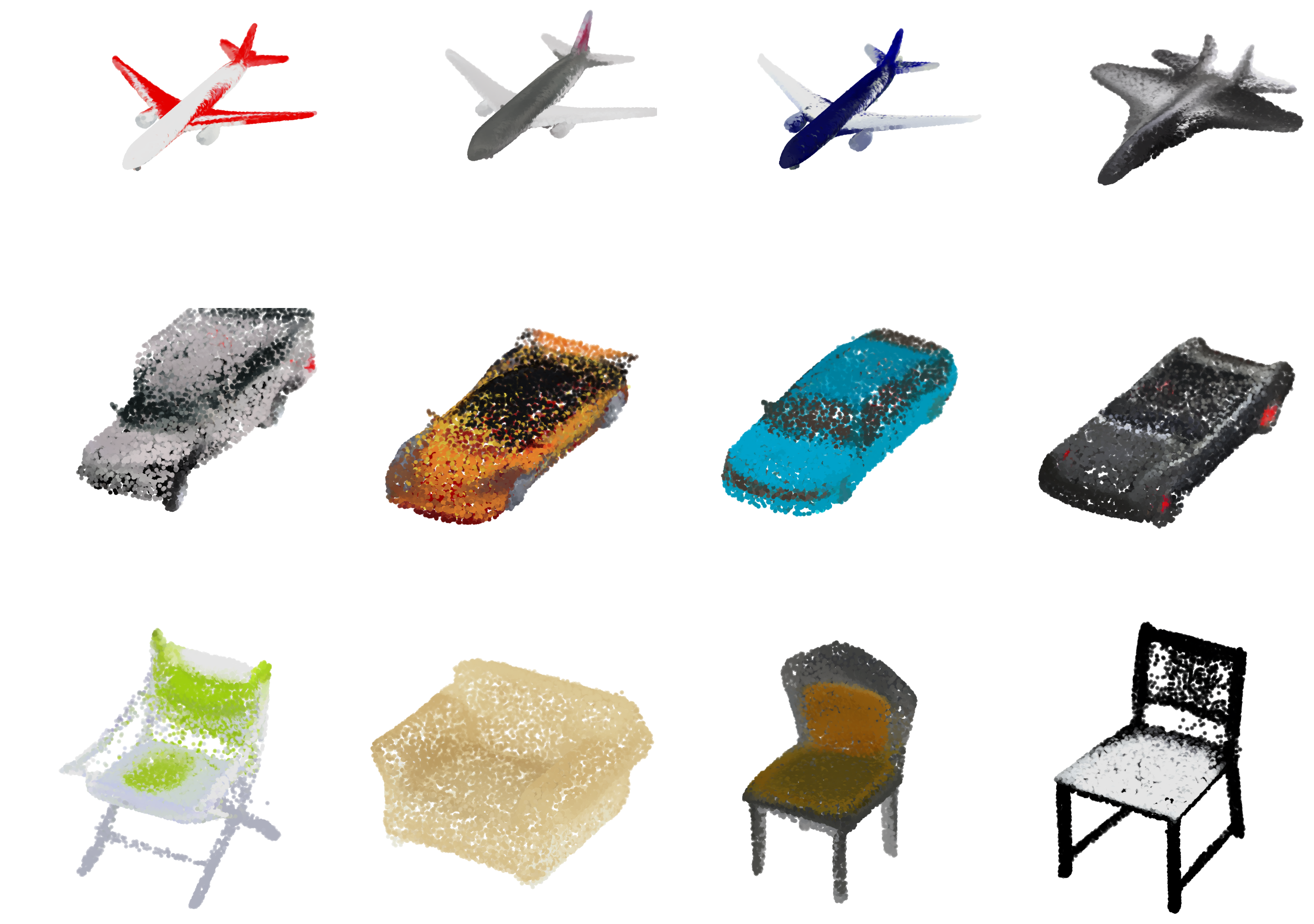}
\end{center} 
\caption{Examples of auto-colored 3D point clouds from three classes of objects generated by \our{} model.}
\label{fig_rec:point:col}
\end{figure*}

Regarding the indoor environments, in \cite{procedural_dungeons} the authors carried out a survey of a content generation method for dungeons and catacombs systems. Another example was given by \cite{indoor_procedural_lvl}, where the authors presented their method of procedural generation of room systems and corridors in 2D/3D games using minimum spanning trees.

\begin{figure}
\begin{center} 
 \includegraphics[width=1\linewidth]{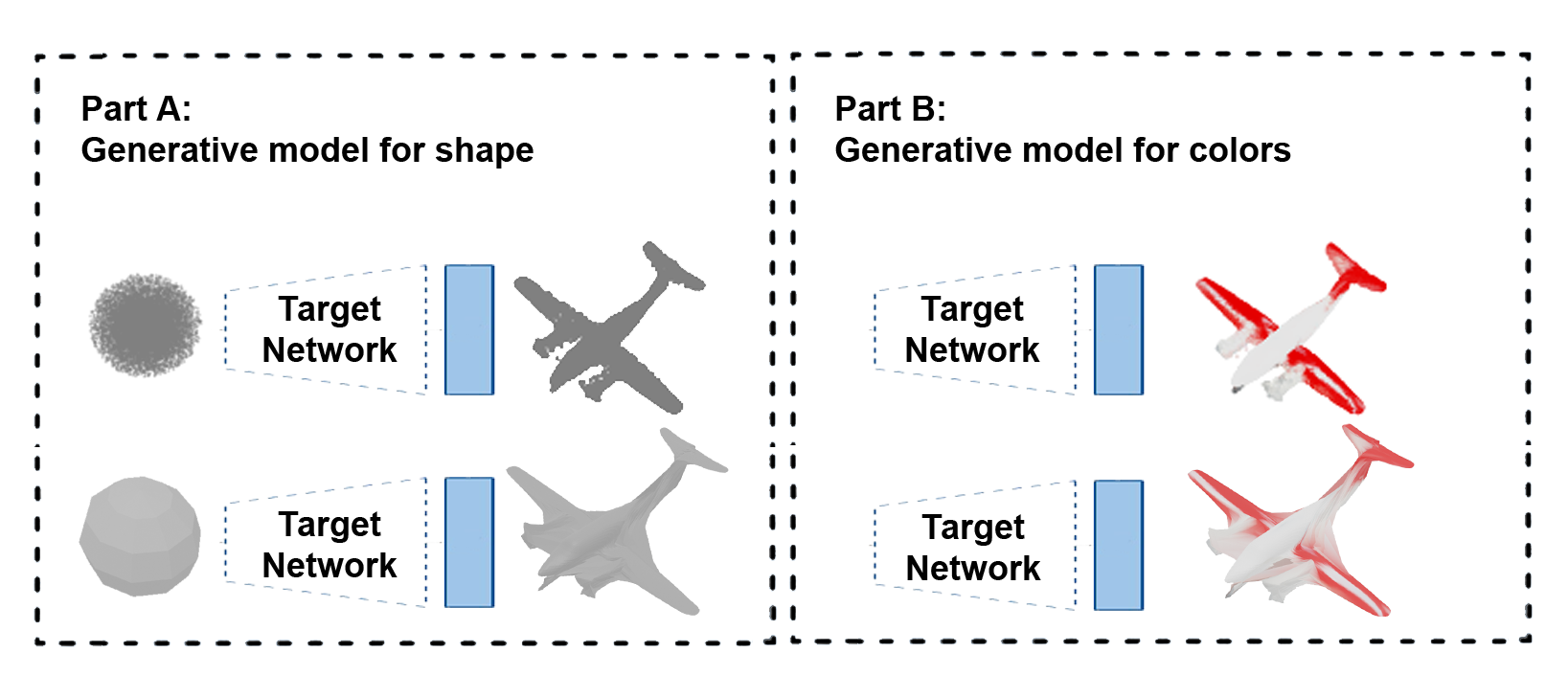}
\end{center} 
\caption{Generation of an auto-colored mesh from a 3D point cloud. The first stage outputs a neural network, which then transfers samples from a uniform distribution on a sphere unit object. Then, the unit sphere is transformed into the border of our data set  (Part A). In the second stage we produce colors for vertices. The final colors on meshes are obtained by interpolation of the colors from vertices (Part B).} 
\label{fig:tt} 
\end{figure}

Moreover, in \cite{8382283}, the authors discuss the advantages that the ML-based approach has over standard procedural methods. For instance, ML is well suited to not only generate various elements of the game levels (e.g. scenes, models, or game mechanics). Due to its statistical nature, it can be also used to analyze the generation outcomes, i.e. the resulting game level and its components. For instance, \cite{GAN_lvl_3D} provides an example of applying Deep Convolutional
Generative Adversarial Networks (DCGANs) to generate educational game levels. The authors remarked that their approach offers higher solvability with a cost of a small decrease in the novelty of game levels.

%Evaluation of the multistep DCGAN model shows that it significantly enhances the solvability of the generated levels with minor degradation in the novelty of the generated levels. 

% \tomek{I don't get this section - is it an intro to the next ones? It seems that it gives an intuition of the method and I would expect subsection of this method to be a detailed version of the ideas mentioned here.}\slawek{Perhaps we could change the title of this section to something like "Auto-colored 3D Object Generation", and then made the next sections its subsections? Here, the "General Idea", would be the first subsection. Also, since we are talking about first-stage, second-stage and triangulation, I would suggest dividing the process with accordance to that description? It would be easier to follow the paper I guess.} 

\section{Auto-colored 3D Object Generation}

In this work, we introduce a two-stage training process that aims at obtaining auto-colored 3D point clouds, which is then followed by applying the {\it triangulation trick} to produce auto-colored 3D meshes. 

\subsection{General idea}

In the first stage, we use an autoencoder to generate 3D meshes of objects. In practice, we can to work with any point on the object's surface. Therefore we use continuous representation of the surface. In the basic pipeline we use HyperCloud \cite{spurek2020hypernetwork}, yet our method is agnostic to a 3D point cloud generative model and can work with other methods, including PointFlow \cite{yang2019pointflow} or HyperFlow \cite{spurek2020hyperflow}. The main idea is to represent a 3D object as a neural network, which transfers uniform distribution on a 3D sphere into the 3D object's surface (see Part A in Fig.~\ref{fig:model}). Thanks to such a solution, we can sample any number of points on the object's surface. Moreover, we can use the {\it triangulation trick} (see Part A in Fig.~\ref{fig:tt}) to generate 3D meshes. %After the second stage, we show that it is possible to modify such a mechanism to produce meshes with colors.
The first stage produces a neural network, which transfers a sample from a uniform distribution on a sphere unit object. In consequence, the unit sphere is transformed into the border of our data set. 

In the second stage, we train the colorization module. To we do this by preparing an additional neural network, which transfers the sample from a uniform distribution into a RGB color space.
The second stage only produces a RGB color for point obtained from the sample (see Part B Fig.~\ref{fig:model}). Our approach allows to produce colors for any point from the surface without changing the position of the reconstructed points. Furthermore, we can combine it with the \textit{triangulation trick} \cite{spurek2020hypernetwork} to produce auto-colored 3D meshes, as we show in Part B in Fig.~\ref{fig:tt}. 

Here, we present a method for generating auto-colored 3D models. First, we introduce generative models that are trained directly on 3D point clouds data. In consequence, we can generate non-colored 3D objects. Next, we extend this pipeline with the module for colorization (see Fig.~\ref{fig_rec:point:col}). As such, our method allows to construct auto-colored 3D meshes of chosen objects (see Fig.~\ref{fig:chair_sphere_triangles}).

\begin{figure*}[!ht]
\begin{center} 
 \includegraphics[width=\linewidth]{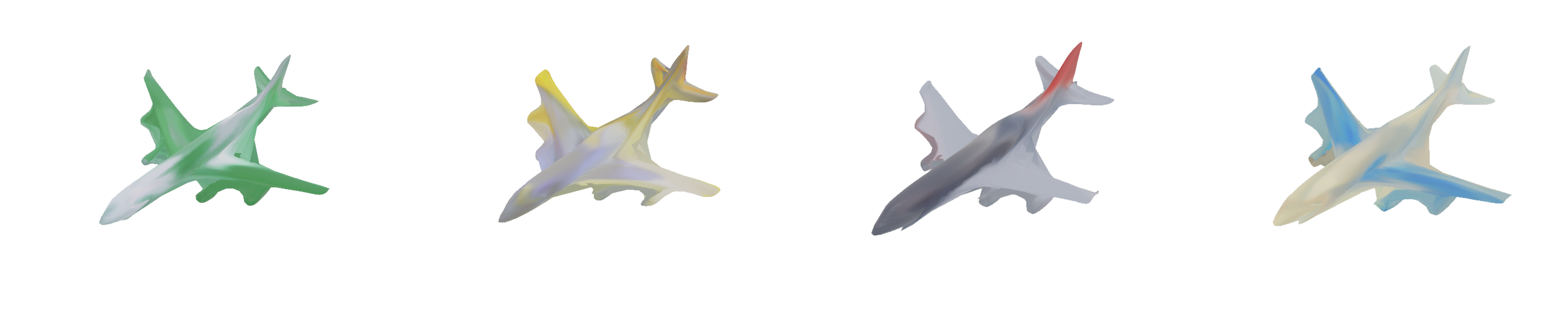}
 \includegraphics[width=\linewidth]{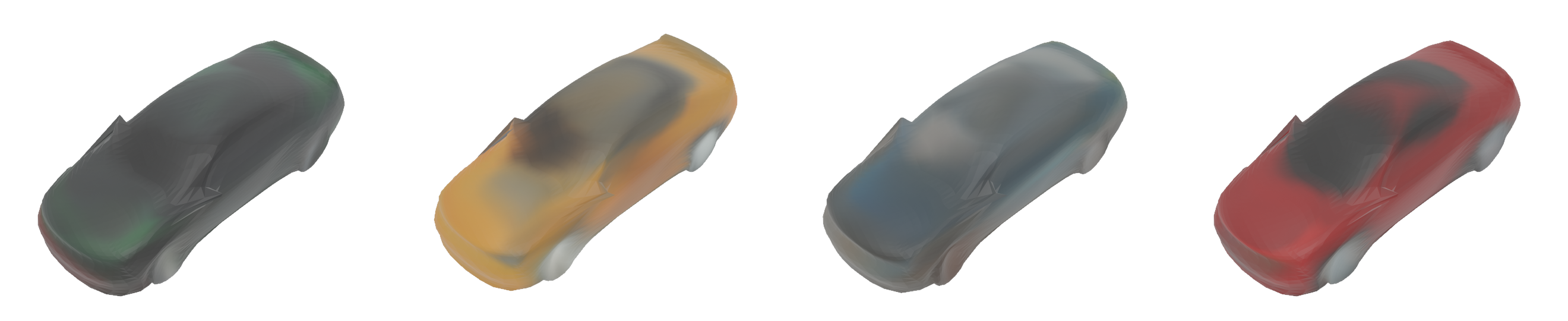}
 \includegraphics[width=\linewidth]{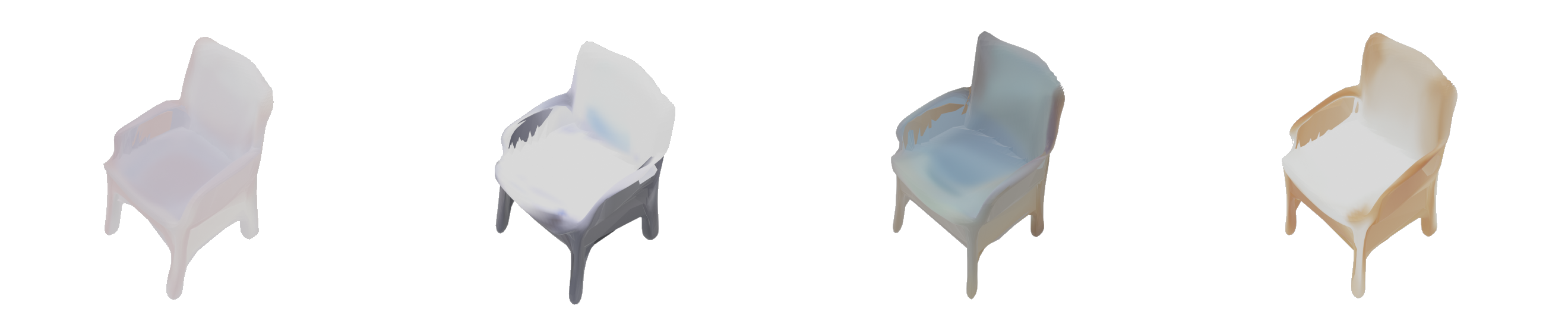}
\end{center} 
%\caption{Our architectures allows to sample many different colors for one reconstruction.}
\caption{Our HyplerCloud-based method allows to sample many different coloring scheme for a single object's reconstruction.}
\label{fig:one_ob_many_colors} 
\end{figure*}

For training, we use $\X \subset \R^6$ where the first three elements encode the position while the last three encode an RGB color. For the first stage, we use only positions. Hence, we denote these Euclidean coordinates as $X_{[0:3]}$.% as the data containing only the Euclidean coordinates.

% In training, we use $\X \subset \R^6$ where triplet is coordinates describe the position and three last encode RGB colors. In the first stage, we use only positions; therefore, we denote by $X_{[0:3]}$ date containing only coordinates of the point without colors.  

\subsection{Generative Models for 3D Point Clouds}
Let us start with the autoencoder architecture for 3D point cloud. Let $X = \{X_i\}_{i=1,\ldots,n} = \{(x_i,y_i,z_i,r_i,g_i,b_i)\}_{i=1,\ldots,n}$  be a given 3D point cloud. Most of existing method works only on coordinates, therefore we will use $X_{[0:3]} = \{X_{[0:3]_i}\}_{i=1,\ldots,n}  = \{(x_i,y_i,z_i)\}_{i=1,\ldots,n}$ 

In general auto-encoder transport the data through latent space $\Z \subseteq \R^D$ and build reconstruction. The architecture consists of an encoder $\E:\X \to \Z$ and decoder $\D:\Z \to \X$, which minimizes the reconstruction error between $X_{[0:3]_i}$ and its reconstructions $\D(\E X_{[0:3]_i})$.

%For 3D point clouds two distance measures are used for reconstruction: Earth Mover’s (Wasserstein) Distance \cite{rubner2000earth} and Chamfer pseudo-distance \cite{tran20133d}.

We use two distance measures for 3D point clouds, namely the Earth Mover’s (Wasserstein) Distance \cite{rubner2000earth} and Chamfer pseudo-distance \cite{tran20133d}.

Earth Mover’s Distance (EMD) \cite{rubner2000earth}  is a metric between two distributions based on the minimal cost that must be paid to transform one distribution into the other. For two equally sized subsets $X_1 \subset \R^3$ and $X_2 \subset \R^3$
their EMD is defined as:
$$
\begin{array}{c}
EMD(X_1,X_2)=\min\limits_{\phi:X_1\to X_2} \sum\limits_{x \in X_1} c(x, \phi(x))
\end{array}
$$

\noindent where $\phi$ is a bijection and $c(x, \phi(x))$ is the cost function that can be defined as: 
$$
\begin{array}{c}
c(x, \phi(x)) = \frac{1}{2} \| x - \phi(x) \|_2^2.
\end{array}
$$

Chamfer pseudo-distance (CD) \cite{tran20133d} measures the squared distance between each point in one set to its nearest neighbor in the other set:
$$
\begin{array}{c}
CD(X_1,X_2)=\!\!\!\! \sum\limits_{x \in X_1} \min\limits_{y\in X_2} \! \| x-y \|_2^2\! + \!\!\!\! \sum\limits_{x \in X_2} \min\limits_{y\in X_1} \! \| x-y \|_2^2.
\end{array}
$$

The task of simultaneous reconstruction of the positions and their colors is non-trivial, as there exists a trade-off between the reconstruction and colorization quality. % We have to solve the trade between reconstruction and colorization. 
Therefore we propose a two-stage strategy to address this issue. First, we produce reconstructions, and then we add colors on top of them. In the experimental section, we compare our model with the one-stage approach that can be considered a baseline (see Tab.~\ref{tab:1}).

Classical auto-encoder can be modified to become a generative model. To that end we change the cost function that forces the model to be generative, i.e. ensures that the data transported to the latent space comes from the prior distribution (typically Gaussian)~\cite{kingma2013auto,tolstikhin2017wasserstein,tabor2018cramer}. Thus, the generative auto-encoder model uses reconstruction loss and the distance of a given sample represented in the latent space from the prior distribution.

In our work, we use Variational Auto-encoders (VAE)~\cite{kingma2013auto}. Such a model uses variational inference \cite{kingma2013auto}  to ensure that the data transported to latent space $\Z$ are distributed according to standard normal density. In practice, we add the distance from standard multivariate normal density:
\begin{equation}
\mathrm{cost}(\X;\E,\D) \! =  \! Err(\X;\D(\E \X )) \!\! + \!\! \lambda D_{KL}(\E \X,N(0,I)),
\label{eq:cost_general}
\end{equation}

\noindent where $ D_{KL}$ is the Kullback–Leibler divergence \cite{kullback1951information}.

% Due to large techniques of enforcing probability distribution on the latent space, the cost function of the model can be formulated in the more general form:
% \begin{equation}
% \mathrm{cost}(\X;\E,\D)= Err(\X;\D(\E \X )) + Reg(\E \X, P),
% \label{eq:cost_general}
% \end{equation}
% where $Err$ is Earth Mover’s (Wasserstein) Distance or Chamfer pseudo-distance and $Reg$ is a function that forces latent space to be from some known or trainable distribution $P$. For known distributions like Gaussian, Kullback–Leibler divergence or adversarial training can be used for regularization. 

\subsection{Hypernetwork}
Hypernetworks~\cite{ha2016hypernetworks} are defined as neural models that generate weights for a separate target network trained to solve a specific task.
% hypernetworks was used to produce continuous representation of 3D objects \cite{spurek2020hypernetwork}. 

Since point clouds contain a number of data points, a method dedicated for such a task must be permutation invariant. In PointNet~\cite{qi2017pointnet} the authors proposed architecture that works with different size of 3D point clouds used as input to the neural network. However, working with varying outputs sizes poses a challenge that PointNet is not designed to solve for. One of the possible solutions is to use a hypernetwork. Instead of producing a fixed-size output, we create a small neural network--called target network--to produce an output of any size understood as a number of points.   

\begin{figure*}
\begin{center} 
 \includegraphics[width=\linewidth]{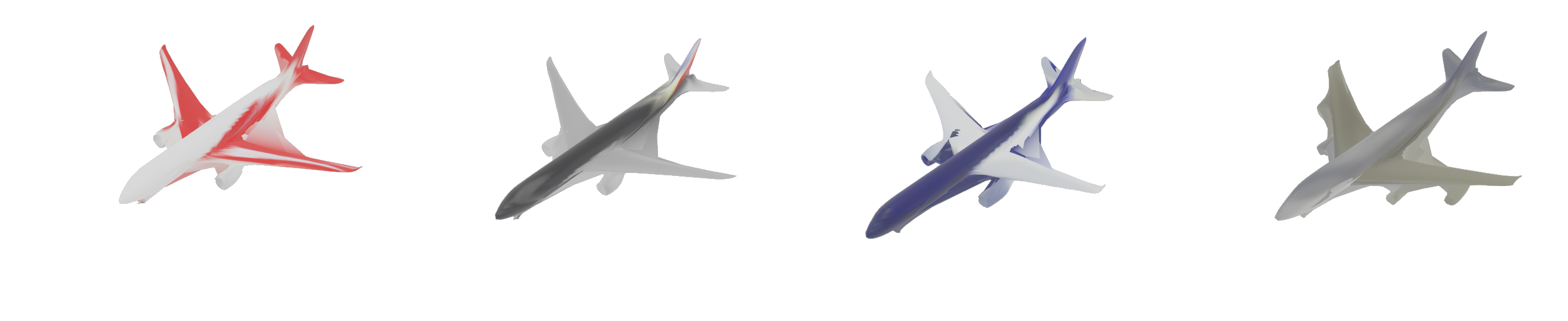}

 \includegraphics[width=\linewidth]{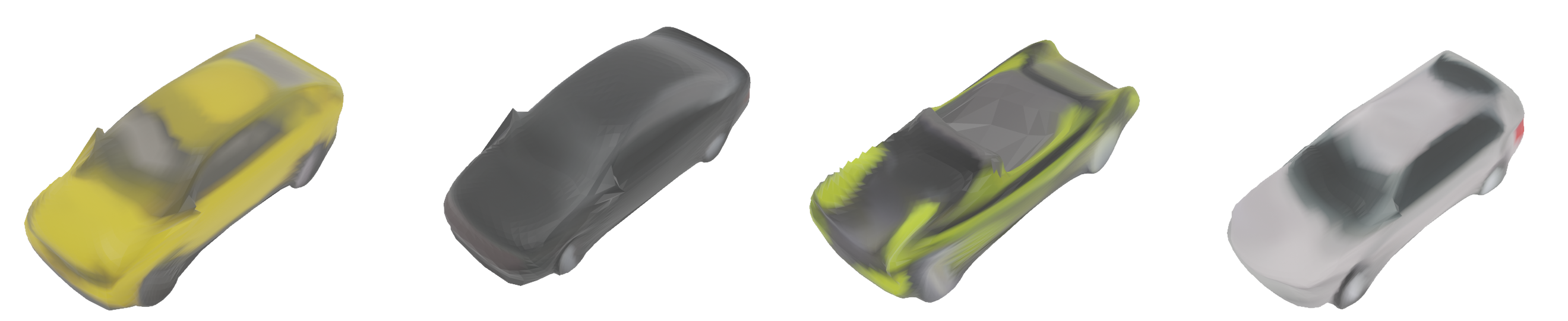}
\includegraphics[width=\linewidth]{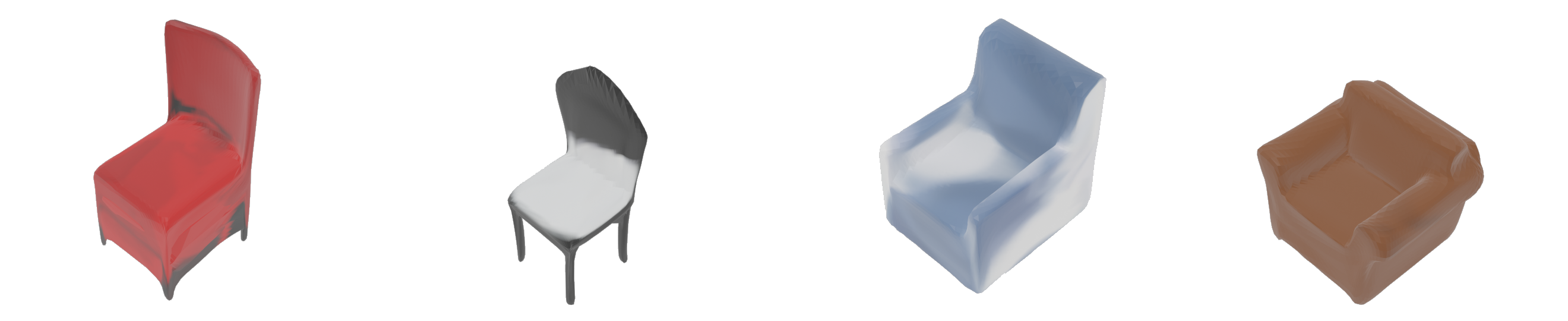}
\end{center} 
\caption{Examples of complex 3D colored meshes generated by our HyperColor.}
\label{fig:chair_sphere_triangles} 
\end{figure*}

The target network is not trained directly.   
We use a hypernetwork that returns weights to the corresponding target network. Only the weights of hypernetwork are trained. 

\subsection{HyperCloud}
In \cite{spurek2020hypernetwork} the authors introduced HyperCloud technique which use hypernetwork to produce continuous representation of 3D objects. Instead of generating object directly with the decoder \cite{zamorski2018adversarial}, the HyperCloud uses parameterization of the 3D object's surface as a function transferring uniform distribution on 3D ball into surface of an object. 

In HyperCloud, instead of producing a 3D point clouds, we generate many neural networks, i.e. we make a different neural network for each object class. In practice, we have one neural network architecture and our model produce different weights. 

More precisely, we model function  $T_{\theta} : \R^3 \to \R^3$ (neural network with weights $\theta$), which takes an element from the prior distribution $\P$ and transfers it on a point on the surface of the object. 

Such an approach allows producing as many points as we need. We can sample an arbitrary number of points from the uniform distribution of the unit ball and transfer them by the target network to the object surface. Furthermore, we can produce a continuous mesh representation of an object by means of the \textit{triangulation trick}. All elements from the ball are transformed into a 3D object. In consequence, the unit sphere is transformed into the surface of the object.

Now we can produce meshes without a secondary rendering procedure. It is obtained by simply feeding our neural network with the vertices of a sphere mesh as shown in Part A in Fig.~\ref{fig:tt}. As a result, we obtain a high-quality 3D meshes. 

We do not train target network directly. We use a hypernetwork 
$
\begin{array}{c}
H_{\phi}: \R^3 \to \theta ,
\end{array}
$
which take an point cloud $X_{[0:3]} \subset \R^3$ and return weights $\theta$ to the target network $T_{\theta}$.
In such framework a point cloud $X_{[0:3]}$ is parametrized by a function 
$$
\begin{array}{c}
T((x,y,x);\theta) = T((x,y,x); H_{\phi}(X)).
\end{array}
$$

Our goal is to train the weights $\phi$ of the hypernetwork. For this purpose, we minimize the distance between point clouds Chamfer distance~\cite{tran20133d}. We take an input point cloud $X_{0:3} \subset \R^3$ and pass it to $H_{\phi}$. The hypernetwork returns weights $\theta$ to the target network $T_{\theta}$. Next, the input point cloud $X_{0:3}$ is compared with the output from the target network $T_{\theta}$. In order to do so we sample the the correct number of points from the prior distribution and transfer them by target network.

% As a hypernetwork, we use a permutation invariant encoder that is based on PointNet architecture \cite{qi2017pointnet} and modified decoder to produce weight instead of row points.
% The architecture of $T_{\theta}$ consists of: an encoder ($\E$) which is a PointNet-like network that transports the data to lower-dimensional latent space $\Z \in \R^D$ and a decoder ($\D$)  (fully-connected
% network), which transfers latent space to the vector of weights for the target network. 
% In our framework hypernetwork $T_{\theta}(X)$ represents our autoencoder structure $\D(\E X)$.
% Assuming  $T_{\theta}(X) = \D(\E X)$, we train our model by minimizing the cost function given by equation (\ref{eq:cost_general}). 

% Observe, that we only train a single neural model (hypernetwork), which allows us to produce a great variety of functions at test time. In consequence, we might expect that target networks for similar point cloud will be similar. We are able to produce smooth interpolation by using hypernetwork.

\subsection{\our{}: Extending Pipeline with Coloring Module}
In the second stage, we add colors to the previously produced 3D point clouds. In turn, this model architecture uses objects generated in the first stage. 
% % Our goal is to add colors to the existing 3D points. 
% In practice, the data processed in the second stage is a copy of the output of the first stage, but its objective is to add colors to the existing 3D points. 
% %working on 3D point clouds with colors to model the distribution of 3D points. 
% On the other hand, in \cite{li2018point} the authors applied a GAN architecture. \tomek{This sentence comes as a surprise to me - why do we talk about some related work here?}All above method works on row 3D point clouds without colors.

We use a two-stage strategy since in a single-stage strategy we have design a method of reconstruction that simultaneously auto-colors the object. In such a framework, we have to solve the trade-offs between reconstruction and colorization. In the experimental section, we compare our model with a single-stage strategy which we consider our baseline.

In our second stage, we model function  $C_{\eta} : \R^3 \to \R^3$ (neural network with weights $\eta$), which takes an element from a uniform distribution and transfers it into the RGB color-space of reconstructed points $T_{\theta}(X_{[0:3]})$\footnote{In practice we use reconstruction in LAB format \cite{gowda2018colornet} and render images in RGB. }.

We use one sample from the 3D uniform distribution and two functions: $T_{\theta}$ and $C_{\eta}$. The first one produces points on the surface, and the second one produces colors for such points. Since we use one sample from the 3D uniform distribution, we share information between shapes and colors.

Analogically to the first stage, we do not train target network $C_{\eta}$ directly.   
We use a hypernetwork
$
\begin{array}{c}
H_{\psi}: \R^3 \supset X \to \eta ,
\end{array}
$
which takes a point cloud with colors $X \subset \R^6$ 
%(three firs coordinates describe coordinates of point and nest three colors describe RGB colors) 
and returns weights $\theta$ to the target network $C_{\eta}$.
The second stage must use data with colors.

In this framework, colors of points $X \subset \R^6$ are described as functions. 
$$
\begin{array}{c}
C((x,y,x,r,g,b);\eta) = C((x,y,x,r,g,b)); H_{\psi}(X)).
\end{array}.
$$

Our goal is to train the weights $\psi$ of the hypernetwork. For this purpose, we minimize the mean square error distance (classical Euclidean distance) between colors of the original object and colors produced by the target network $C_{\eta}$.

We take an input point cloud $X_{[0:3]} \subset \R^3$ and pass it to $H_{\phi}$.  The hypernetwork returns weights $\theta$ to the target network $T_{\theta}$. Next, we obtain a reconstruction of the input object produced by the target network $T_{\theta}$ (we sample the correct number of points from the prior distribution and transfer them by target network to the object surface).

Then, we take the point cloud $X \subset \R^6$ and pass it to $H_{\psi}$. The hypernetwork returns weights $\eta$ to target network $C_{\eta}$ which in turn, produces colors for the reconstructed point cloud. Since we have a position from the first stage, we can use the MSE:
$$
MSE(A,B) = \frac{1}{n} \sum_{i=1}^n (a_i - b_i)^2,
$$
 to train the second stage (colors). The second stage does not have to be invariant to permutations. We simply add colors to each point separately.

% \tomek{This description asks for some sort of a diagram or maybe an Algorithm block?} \slawek{Second this, a pseudo-code diagram may be a really neat addition here.}
Our full procedure of reconstruction consist of four high-level steps that further split into sub-steps where appropriate:
\begin{enumerate}
    \item Take 3D point cloud $X \subset \R^6$. 
    \item Sample $S$ from a uniform distribution on sphere. (Here, we draw as many points as we want to reconstruct.)
    \item Reconstruct shape:
    %\begin{itemize}
    \begin{enumerate}
        \item Pass $X_{[0:3]}$ to hypernetwork $H_{\phi}$.
        \item $H_{\phi}$ returns weights to the target network $T_{\theta}$.
        \item Transfer $S$ by target network $T_{\theta}$ to produce object reconstruction.
    \end{enumerate}
    \item Reconstruct colors:
    \begin{enumerate}
        \item Pass $X$ to hypernetwork $H_{\psi}$.
        \item $H_{\psi}$ returns weights to the target network $C_{\eta}$.
        \item Transfer $S$ by target network $C_{\eta}$ to produce colors for object reconstruction.
    \end{enumerate}    
\end{enumerate}

% \begin{algorithm}[h!]
% \caption{\slawek{A listing for pseudo code of our method}}
% \label{alg:DetaSetGenerator}
% \begin{algorithmic}[1]
% \Require
% \Ensure
% \\
% \Procedure{Generate3DPointCloud}{}\\
% \Comment{A first stage of our method.}
% \EndProcedure
% \\
% \Procedure{Color3DPointCloud}{}\\
% \Comment{A second stage of our method.}
% \EndProcedure
% \\
% \Procedure{GenerateColored3DMesh}{}\\
% \Comment{Generating 3D mesh out of colored point cloud.}
% \EndProcedure
% \end{algorithmic}
% \end{algorithm}

Once we have two point clouds, i.e. the original one and the reconstructed one produced in our two-stage process, we need to find the alignment between individual points. To that end, we find the closest element from the original object for each point from the reconstructed point cloud. We use a k-nearest neighbor \cite{zhang2007ml} algorithm for such a task. 

To produce colored meshes from colored 3D point cloud we use the {\it triangulation trick} (see Fig.~\ref{fig:tt}). First, we train a neural network, which transfers samples from a uniform distribution on sphere into object. In consequence, the unit sphere is transformed into the border of our data set. Second stage produce colors for vertex from the produced mesh. The final coloring scheme is obtained by interpolation of colors from vertex (see Part B in Fig.~\ref{fig:tt}).

In order to visualize a colored mesh we use Blender's Vertex Paint module \cite{blender_vertex_paint}. It allows us to directly set the color of vertices instead of using textures. It results in faces having a gradient calculated using its assigned vertex colors. Then we can obtain a textured object through the process of baking vertex color into an image.

\begin{figure*}
\begin{center} 
 \includegraphics[width=\linewidth]{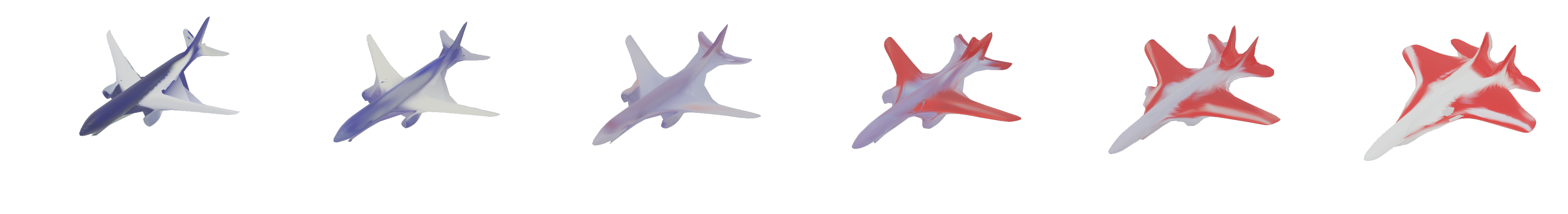}
 \includegraphics[width=\linewidth]{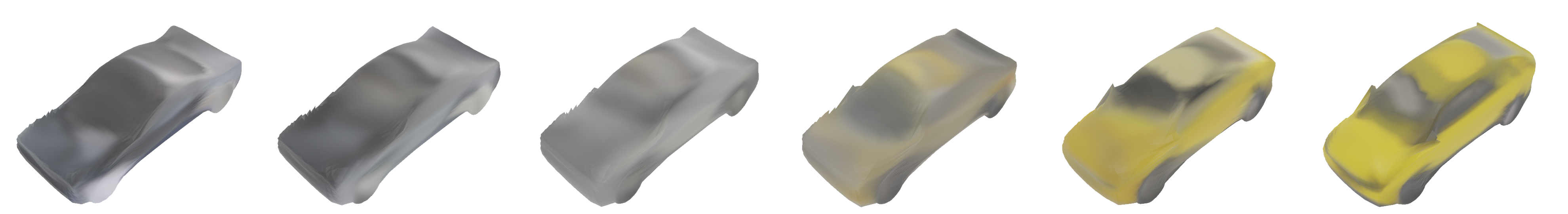}
 \includegraphics[width=\linewidth]{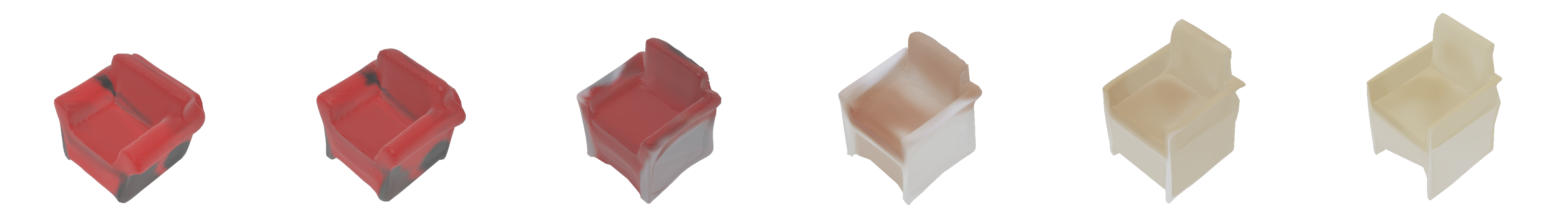}
\end{center} 
\caption{Example of interpolation between two objects with different colors. As we can see, we have smooth changes in shape as well as in color space. }
\label{fig:chair_sphere_triangles_interpolation} 
\end{figure*}

%\section{Experiments}
\section{Results and Discussion}
In this section, we present the experimental results of the proposed generative models in generating auto-colored point clouds and meshes. Here, we report the results indicating that our model gives a better quality of shape reconstruction and coloring in comparison to a single-stage strategy. Next, we show that we obtain a generative model, which produces colored meshes.

\subsection{3D Point Cloud Reconstruction Capabilities}
We now evaluate how well our model encodes a point cloud. We conducted an autoencoding task for 3D point clouds from three categories from ShapeNet dataset (\textit{airplane}, \textit{car}, \textit{chair}). Next, we evaluated the Chamfer distance between original shapes and reconstruction obtained in the first stage. We also gave the MSE measure between original colors and the reconstructed ones. As can be seen in Tab.~\ref{tab:1}, our double-stage strategy gives better reconstruction in shape as well as in color space. As we can see, the two-stage strategy allows us to produce colors without any loss of reconstruction quality. Examples of auto-colored point clouds produced by our model can be seen in Fig.~\ref{fig_rec:point:col}.

\begin{table}[!h]
\caption{Evaluation of our approach and the baseline methods in the reconstruction task. As we can see, our two-stage strategy gives essentially better results.}
\begin{center} 
\scalebox{0.91}{
    \begin{tabular}{ |l||c|c|  }
        \hline
        Object & \multicolumn{2}{|c|}{Airplane } \\
        \hline
        Measure & Chamfer - shape & MSE - colors \\
        \hline
        Base & $3.173 \times 10^{-4} \pm 4.326 \times 10^{-4}$ & $1977.461 \pm 7189.232$ \\
        \hline
        \our{} &  $\bf 4.053 \times 10^{-5} \pm 6.934 \times 10^{-5}$ &  $\bf 898.079 \pm 1737.289$ \\
        \hline
        
        \hline
        Object & \multicolumn{2}{|c|}{Car } \\
        \hline
        Measure & Chamfer - shape & MSE - colors \\
        \hline
        Base & $7.211 \times 10^{-4} \pm 7.015 \times 10^{-4}$ & $3387.089 \pm 8210.053$ \\
        \hline
        \our{} &  $\bf 1.260 \times 10^{-4} \pm 4.291 \times 10^{-5}$ &  $\bf 1541.382\pm 2418.891$ \\
        \hline
        
        \hline
        Object & \multicolumn{2}{|c|}{Chair } \\
        \hline
        Measure & Chamfer - shape & MSE - colors \\
        \hline
        Base & $5.476 \times 10^{-4} \pm 2.951 \times 10^{-4}$ & $2087.659 \pm 7292.511$ \\
        \hline
        \our{} &   $ \bf 1.074 \times  10^{-4} \pm 3.828 \times  10^{-5}$ & $ \bf 359.996 \pm 1083.338$ \\
        \hline
    \end{tabular}
% \end{center}
% \end{small}
}
\end{center}

\label{tab:1}
\end{table}

%%%%%%%%%%%%%%%%%%%%%%%%%%%%%%%%%%%%
%\paragraph{Generation of 3D meshes}
\subsection{Generation of 3D Meshes}
%%%%%%%%%%%%%%%%%%%%%%%%%%%%%%%%%%%%
The main advantage of our method is the ability to generate both auto-colored 3D point clouds as well as auto-colored 3D meshes. We achieve this without any need for post-processing. In Fig.~\ref{fig:chair_sphere_triangles}, we present auto-colored meshes. Thanks to using a uniform distribution on the 3D ball, we can easily construct a mesh of an object. All elements from the ball are transformed into a 3D object. In consequence, the unit sphere is transformed into surface of the object. As it was mentioned, we can produce meshes without a secondary meshing procedure. It is obtained by propagating the triangulation of the 3D sphere through the target network as shown in Fig~\ref{fig:tt}. 

When obtained a 3D mesh we can color its vertices in the second-stage using similar procedure as in the first-stage. Next, we can produce colored meshes by simple interpolation.

Our model is bale to produce many different colors for single object (see Fig.~\ref{fig:one_ob_many_colors}).

%\paragraph{Interpolation}
\subsection{Interpolation}
In our model, we can construct interpolation between objects. We have two prior distributions: Gaussian in the first and second stages. Therefore, we can produce smooth transition in shape and color space, see Fig.~\ref{fig:chair_sphere_triangles_interpolation}.

% \begin{figure*}
% \begin{center} 
%  \includegraphics[width=0.75\linewidth]{img/color_3.png}
% \end{center} 
% \caption{Architecture.} 
% \label{fig:color_3} 
% \end{figure*}

% \begin{figure*}
% \begin{center} 
%  \includegraphics[width=\linewidth]{img/game_scene_chairs.png}
% \end{center} 
% \caption{Chairs in a game scene.}
% \label{fig:wizualization} 
% \end{figure*}

% \begin{figure*}[h!]
% \begin{center} 
%  \includegraphics[width=\linewidth]{img/game_scene_chairs_selected.png}
% \end{center} 
% \caption{Chairs in a game scene. The scene background was constructed out of photogrammetric model of \textit{the Great Drawing Room} at The Hallwyl Museum downloaded from \cite{Maggiordomo2020RealWorldTT}}
% \label{fig:wizualization} 
% \end{figure*}

% \section{Discussion}
% \slawek{Tutaj poki co nic nie mamy? Na te chwile nie mam pomyslu co tutaj mozna wpisac. Moze zrobic "merge" z poprzednia sekcja? Cos w stylu: Results and Discussion?}\tomek{Zgadzam się}

\section{Conclusion}
In this work we presented a hypernetwork \cite{spurek2020hypernetwork} approach to synthesizing 3D models of selected class of real-life objects. Using our approach we can swiftly generate any quantity of diversified 3D models used for populating a given game scene. The conduced experiments suggest that our two-stage approach gives better results in terms of shape reconstruction and coloring as compared to traditional single-stage techniques (see Tab.~\ref{tab:1}).

Furthermore, the unique attribute in our method is the automatic and coloring of the 3D models generated this way Fig.~\ref{fig_rec:point:col}--\ref{fig:gamescenesPlane}. As Fig.~\ref{fig:gamescenesChairsCars}--\ref{fig:gamescenesPlane} shows, such models can be easily embedded within the game scene background to compose plausible final effect.

\section{Future Work}
In the future, we plan to advance our method to produce 3D models with finer details, i.e. with more complex meshes including clear coloring borderlines.

Another interesting avenue of research is automatic placement of the models in the game scene. At that point, the algorithm is required to consider the correct orientation of the model with respect to the scene's topology as well as the prior placement of other models. 

Moreover, since in the context of video games the most important factors is the player's satisfaction, we plan to carry out controlled experiments in order to assess with the users how they perceive game scenes populated with models generated with our method.

%In order to achieve this, we will run a comparative study with a large enough number or participant to carry out an ANOVA testing procedure.

% or


% \appendices

% % you can choose not to have a title for an appendix
% % if you want by leaving the argument blank
% \section{}

% % use section* for acknowledgment
% \ifCLASSOPTIONcompsoc
%   % The Computer Society usually uses the plural form
%   \section*{Acknowledgments}
% \else
%   % regular IEEE prefers the singular form
\section*{Acknowledgment}
This research was funded by the Priority Research Area Digiworld under the program Excellence Initiative –Research University at the Jagiellonian University in Kraków.
% \fi


\ifCLASSOPTIONcaptionsoff
  \newpage
\fi

\end{document}